\pdfoutput=1

\documentclass[11pt]{article}

\usepackage{EMNLP2023}

\usepackage{times}
\usepackage{latexsym}

\usepackage[T1]{fontenc}

\usepackage[utf8]{inputenc}

\usepackage{microtype}

\usepackage{inconsolata}
\usepackage{natbib}
\usepackage{csquotes}
\usepackage{graphicx}
\usepackage{amsmath}
\usepackage{amsthm}
\usepackage{amssymb}
\usepackage{algorithm}
\usepackage{algorithmic}
\usepackage{multirow}
\usepackage{colortbl}
\definecolor{mygray}{gray}{.9}
\usepackage{booktabs}
\usepackage{bbm}
\usepackage{subfigure}
\usepackage{hyperref}
\usepackage{array}
\usepackage{makecell}

\newcommand{\Mobius}{\emph{M\"{o}bius\,addition}}
\newcommand{\Mobiustimes}{\emph{M\"{o}bius\,matrix\,multiplication}}
\graphicspath{{./}}

\title{PE: A Poincare Explanation Method for Fast Text Hierarchy Generation}

\author{Qian Chen$^1$, Dongyang Li$^1$, Xiaofeng He$^{1,2}$
, Hongzhao Li$^3$, Hongyu Yi$^3$ \\
  $^1$ School of Computer Science and Technology, East China Normal University, Shanghai, China 
  \\ 
  $^2$ NPPA Key Laboratory of Publishing Integration Development, ECNUP, Shanghai, China \\
  $^3$ Sichuan Caizi Software Information Network Co., Ltd\\
  \texttt{\{qianchen901005,fromdongyang\}@gmail.com,}  \texttt{hexf@cs.ecnu.edu.cn,}
  \texttt{lhz@sxw.cn,}\texttt{alwayslater@yeah.net}\\}

\begin{document}
\maketitle
\begin{abstract}
The black-box nature of deep learning models in NLP hinders their widespread application.  The research focus has shifted to Hierarchical Attribution (HA) for its ability to model feature interactions. Recent works model  non-contiguous combinations with a time-costly greedy search in Eculidean spaces, neglecting  underlying linguistic information in feature representations. In this work, we introduce a novel method, namely Poincare Explanation (PE), for modeling feature interactions  with hyperbolic spaces in a time efficient manner.
Specifically, we take building text hierarchies as finding spanning trees in hyperbolic spaces. 
First we project the embeddings into hyperbolic spaces to elicit inherit semantic and  syntax hierarchical structures. 
Then we propose a simple yet effective strategy to calculate Shapley score. Finally we build the the hierarchy with proving the constructing process in the projected space could be viewed as building a minimum spanning tree and introduce a time efficient building algorithm. Experimental results demonstrate the effectiveness of our approach. 
\end{abstract}

\section{Introduction}
Deep learning models have been ubiquitous in Natural Language Processing (NLP) areas accompanied by the explosion of the parameters, leading to increased opaqueness.  Consequently, a series of interpretability studies have emerged \cite{attention_flow, transformers_key_value_memory, unfied_peft}, among them feature attribution methods stand out owing to fidelity and loyalty axioms and straightforward applicability \cite{feature_survey}.

Previous feature-based works are limited to single words or phrases \cite{ig1}.  However,  \citet{analysis_lime} point out that LIME's \cite{lime} performance on simple models is not plausible \footnote{A figure illustration is provided in Appendix \ref{lime_expl}}. To model feature interactions, Hierarchical Attribution (HA) \cite{chen-etal-2020-generating-hierarchical,he} has been introduced, with a attribution-then-cluster stage in which constructs feature interaction process by distributing text group scores at different levels\footnote{A vivid HA example is provided in Appendix \ref{HA example}.}.
From bottom to the up, HA categorizes all words into different clusters,  ending with a tree structure.

\begin{figure}[!tb]
    \centering
    \includegraphics[width=0.5\textwidth]{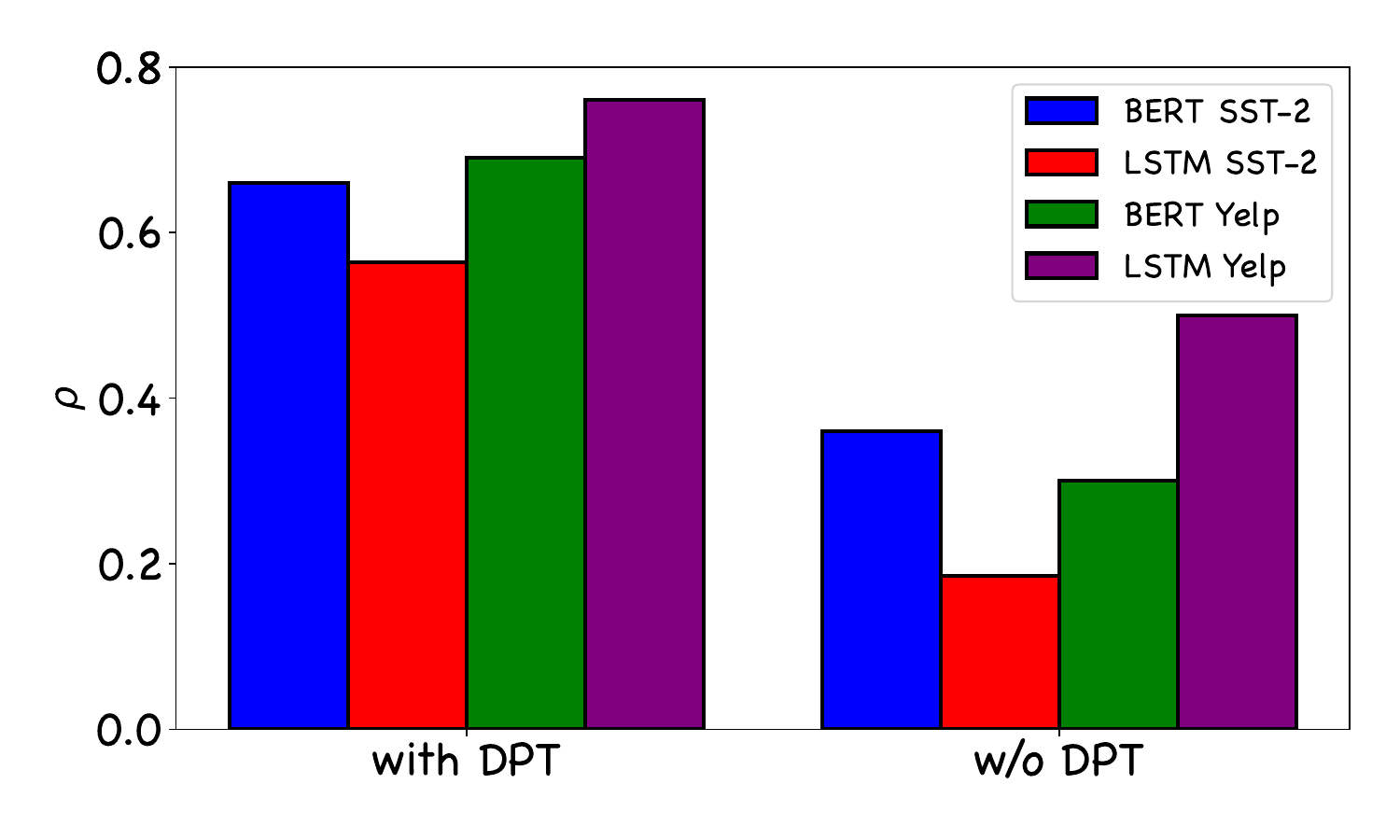}
    \caption{Pearson correlation $\rho$ results from \citet{ig_v1} with BERT and LSTM on SST-2 and Yelp datasets. A higher correlation coefficient indicates a stronger ability of the method to identify important words.}
    \label{fig:intro2}
\end{figure}

\begin{figure}[!tb]
    \centering
    \includegraphics[width=0.45\textwidth]{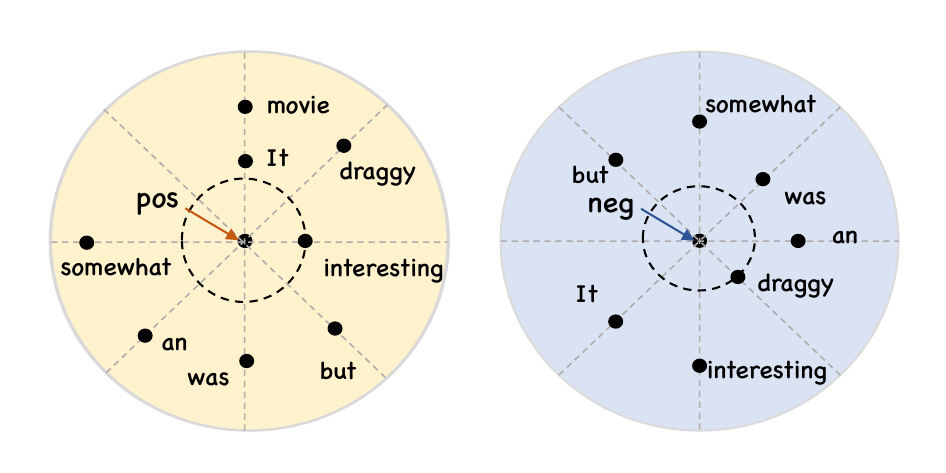}
    \caption{Left: The projection illustration for positive example \enquote{\emph{It was an interesting but somewhat draggy movie}.} The centre represents the prototype for the positive label. Right: A negative example \enquote{\emph{It was a draggy but somewhat interesting movie}.} The center point stands for the negative label.}
    \label{fig:model1}
\end{figure}

However, building feature hierarchies is not a trivial thing. Existing methods have three following problems. \textbf{P-1}: Detecting contiguous text spans to replace all possible interactions \cite{singh2018hierarchical,chen-etal-2020-generating-hierarchical}. Only using  spans might lose long-range dependencies in text \cite{transformer}. For example, in the positive example \enquote{\emph{Even in moments of sorrow, certain memories can evoke happiness}}, (\enquote{\emph{Even}}, \enquote{\emph{sorrow}}) is vital and non-adjacent.  \textbf{P-2}: Current algorithms estimating the importance of feature combinations are accompanied by lengthy optimization processes \cite{he,chen-etal-2020-generating-hierarchical}. For example, HE \cite{he} estimates the importance of words using  LIME algorithm and then enumerates word combinations to construct the hierarchy, with a cubic time complexity\footnote{For  convenience of comparison, we ignore the time taken by linear regression  in  LIME algorithm and detailed discussion is in Section \ref{discussion}.}. ASIV \cite{asiv}  uses directional Shapley value to model the direction of feature interactions, while  estimating Shapley value requires exponential time.
\textbf{P-3}: Previous methods cannot model the linguistic information including syntax and semantic information. Syntax and semantics can help to construct a hierarchical tree. For syntax, \citet{ig_v1} build hierarchies directly on Dependency Parsing Trees (DPT) and compute  Pearson Correlation (i.e.$\rho$). The results in Figure \ref{fig:intro2} demonstrate syntax could contribute to building explainable hierarchies by reaching a higher correlation.
For semantic, 
we take Figure  \ref{fig:model1} as an example, the hierarchy in hyperbolic space has already achieved preliminary interpretability with the proximity corresponding the polarity.

As the input text length continues to increase, 
efficiently modeling the interaction of non-contiguous features has become a key challenge in promoting HA.
Building a hierarchical attribution tree based on the input text is essentially a \emph{hierarchical clustering} problem. The definition is as follows: given words and their pairwise similarities, the goal is to construct a hierarchy over clusters (word groups). PE approaches this problem by following three steps. First, to model linguistic hierarchical information, we project word embeddings into hyperbolic spaces to uncover hidden semantics and syntax structures. Next, inspired by cooperative game theory \cite{cooperative_game_theory}, we regard  words as players and clusters as coalitions and introduce a simple yet effective strategy to estimate the Shapley score contribution. Finally  we calculate pairwise similarities and propose an algorithm that conceptualizes the bottom-up clustering process  as generating a minimum spanning tree.

Our contributions are summarized as follows:
\begin{itemize}
    \item We propose a method, PE,  using hyperbolic geometry for generating  hierarchical explanations, revealing the feature interaction process. 
    \item PE introduces a fast algorithm for generating hierarchical attribution trees that model non-contiguous feature interactions.
    \item We evaluate the proposed method on three  datasets with BERT \cite{bert}, and the results demonstrate the effectiveness.
\end{itemize}

\section{Related Work}
Feature importance explanation methods mainly assign attribution scores to features \cite{att_token,mix_context,decompx}. Methods can be classified into two categories: single-feature explanation type and multi-feature explanation type.
\subsection{Single-Feature Explanation}
Earlier researches focus on single feature attribution \cite{lime,sundararajan2017axiomatic,shapley}. For example, LIME \cite{lime} aims to fit the local area of the model by linear regression with sampled data points ending with linear weights as attribution scores. Gradient\&Input (Grad$\times$Inp) \cite{grad*shap} combines the gradient norm with Shapley value \cite{shapley1953value}. Deeplift \cite{deeplift} depends on activation difference to calculate attribution scores. IG  \cite{sundararajan2017axiomatic,DIG,enguehard-2023-sequential} uses path integral to compute the contribution of the single feature to the output. It is noticeable that IG is the unique path method to satisfy the completeness and symmetry-preserving axioms. There exist several variants of IG. DIG \cite{DIG} regards similar words as  interpolation points to estimate the integrated gradients value. SIG \cite{enguehard-2023-sequential}
computes the importance of each word in a sentence while keeping all other words fixed. However, scoring individual features is incompatible with interactions between features.

\subsection{Multi-Feature Explanation}
Multi-feature explanation methods aim to model feature interactions in deep learning architectures. For example, \citet{shapley-taylor-index} proposes a variant of Shapley value to measure the interactions. \citet{zhangquanshi-shapley} defines the multivariant Shapley value to analyze interactions between two sets of players. \citet{sian} proposes a sparse interaction additive network to select feature groups. \citet{archipelgo} proposes  an Archipelago framework to measure feature attribution and  interaction  through ArchAttribute and ArchDetect. \citet{asiv} proposes ASIV to model asymmetric higher-order feature interactions. To illustrate the feature interplay process completely, the explanation of feature interaction could be articulated within a hierarchical framework. HEDEG \cite{chen-etal-2020-generating-hierarchical} designs a top-down model-agnostic hierarchical explanation method, with neglecting non-contiguous interactions. \citet{he} addresses the connecting rule limitation in HEDGE, and proposes a greedy algorithm , HE, for generating hierarchical explanations, which is time-costly. And they all neglect lingustice information including syntax and semantics.

\section{Background}

We first give a review of hyperbolic geometry. 

\noindent\textbf{Poincare ball} A common representation model in hyperbolic space is the Poincare ball, denoted as $(\mathcal{B}_c^m, g_{\boldsymbol{x}}^{\mathcal{B}})$, where $c$ is a constant greater than $0$. $\mathcal{B}_c^m = \{\boldsymbol{x} \in \mathbb{R}^m \mid c \left\|\boldsymbol{x}\right\|^2 < 1\}$ is a Riemannian manifold, and $g_{\boldsymbol{x}}^{\mathcal{B}} = (\lambda_{\boldsymbol{x}}^c)^2 \boldsymbol{I}_m$ is its metric tensor, $\lambda_{\boldsymbol{x}}^c=2/(1-c\left\|\boldsymbol{x}\right\|^2)$ is the conformal factor and $c$ is the negative curvature of the hyperbolic space. PE uses the standard Poincare ball with $c=1$. The distance for $\boldsymbol{x},\boldsymbol{y}\in \mathcal{B}_c^m$ is:\\
\begin{equation}
    d_{\mathcal{B}}(\boldsymbol{x},\boldsymbol{y})=2\tanh^{-1}{\left\|-\boldsymbol{x}\oplus_{c}\boldsymbol{y}\right\|},
\end{equation}
where $\oplus_c$ denotes the \Mobius. 
We use $\otimes_c$ to denote the \Mobiustimes. The \Mobius\,for $\boldsymbol{x}$, $\boldsymbol{y}\in\mathbb{R}^m$ is defined as \cite{mobouis}:
\begin{equation}
    \boldsymbol{x}\oplus_c\boldsymbol{y}=\frac{(1+2\langle\boldsymbol{x},\boldsymbol{y}\rangle+\Vert\boldsymbol{y}\Vert^2)\boldsymbol{x}+(1-\Vert\boldsymbol{x}\Vert^2)\boldsymbol{y}}{1+2\langle\boldsymbol{x},\boldsymbol{y}\rangle+\Vert\boldsymbol{x}\Vert^2\Vert\boldsymbol{y}\Vert^2}.
\end{equation}
Given a linear projection $\boldsymbol{A}:\mathbb{R}^m\rightarrow\mathbb{R}^{p}$ and $\boldsymbol{x}\in \mathcal{B}_c^m$, 
then the \Mobiustimes is defined as \cite{mobouis}:
\begin{equation}
    \begin{split}
        &\boldsymbol{A}\otimes_c\boldsymbol{x}=\tanh(\frac{\Vert\boldsymbol{Ax}\Vert}{\Vert\boldsymbol{x}\Vert}\tanh^{-1}(\Vert\boldsymbol{x}\Vert))\frac{\boldsymbol{Ax}}{\Vert\boldsymbol{Ax}\Vert}.
    \end{split}
\end{equation}
\noindent\textbf{Cooperative  Game Theory} We use $N$ to denote a set of players (i.e. token set). A game is a pair $\Gamma=(N,v)$ and $v:2^N\rightarrow \mathbb{R}$ is the characteristic function. A coalition is any subset of $N$. In a cooperative game, players can form coalitions, and each coalition $S\subseteq N$ has a value $v(S)$. 
\section{Methodology}

This section  provides a detailed introduction to the three parts of PE.  First, we need to score each feature; then, based on these scores, we construct a hierarchy. In Section \ref{poincare_projection}, we consider semantic and syntax factors. Besides we facilitate feature Shapley contribution calculation in Section \ref{feature_contribution}. In Section \ref{minimum_spanning_tree}, we combine these factors to score each feature and  propose a fast algorithm for constructing the hierarchy.

\subsection{Poincare Projection}
\label{poincare_projection}
In this paper, we choose Probing \cite{structural_probe} to recover information from embeddings. Namely, we train two matrices to project the Eculidean embeddings to hyperbolic spaces. For a classification task, given a sequence $X_i=\{x_j\}_{1\leq j\leq n}$ and  a trained model $f$, $n$ is the sequence length.
$\hat{y}$ represents the predicted label, and $f(\cdot)$ represents the model's output probability for the predicted label. 
\subsubsection{Label Aware Semantic Probing}
\label{label_aware_probing}
In this subsection, we extract the semantics from the embeddings through probing. We project the embeddings into a hyperbolic space using a transformation matrix. In this space, the distribution of examples with different semantics will change according to their semantic variations. 
First, we feed the sequence $X_i$ into a pre-trained language model to obtain the  contextualized representations $\boldsymbol{E}_i\in\mathbb{R}^{n\times d_{in}}$, with $d_{in}$ denotes the output dim. Next, the sentence embedding $\boldsymbol{s}_i\in\mathbb{R}^{d_{in}}$ is obtained by the hidden representations of the special tag (e.g.[CLS]), which is the first token of the sequence and  used for classification tasks. Our probing matrix consists of two types: $\boldsymbol{A}_{se}, \boldsymbol{A}_{sy}\in\mathbb{R}^{d_{in}\times d_{out}}$ ($d_{out}$ denotes the projection dim) for probing label-aware semantic information and   syntax information. For semantics, we can obtain the projected representation:
\begin{equation}
    \boldsymbol{s}_{i}^{se} = \boldsymbol{A}_{se}\otimes_c \boldsymbol{s}_i.
\end{equation}
Also we can obtain the token presentation:
\begin{equation}
    \boldsymbol{e}_j^{se}=\boldsymbol{A}_{se}\otimes_c\boldsymbol{e}_j.
\end{equation}
To train the probing matrices, we draw inspiration from prototype networks \cite{prototype_nn}, assuming that there exist 
$k$ centroids representing labels in the hyperbolic space. The closer a point is to a centroid, the higher the probability that it belongs to that category. Specifically, instead of using mean pooling to calculate the prototypes, we directly initialize the prototype embeddings in hyperbolic space, denoted as  $\boldsymbol{\omega}=\{\boldsymbol{c}_k\}$ ($\boldsymbol{c}_k$ is the $k$-th label centroid). Given a distance  $d_{\mathcal{B}}$, the prototypes produce a distribution over classes for a point $\boldsymbol{x}$ based on a softmax over distances to prototypes in the embedding space:
\begin{equation}
    \mathcal{P}(y=k\mid \boldsymbol{\omega})=\frac{\exp(-d_{\mathcal{B}}(\boldsymbol{s}_{i}^{se},\boldsymbol{c}_k))}{\sum_{k^{\prime}}{\exp(-d_{\mathcal{B}}(\boldsymbol{s}_{i}^{se},\boldsymbol{c}_{k^{\prime}}))}}.
\end{equation}
We minimize the negative log-probability $J(\boldsymbol{\omega})=-log\mathcal{P}(y=k\mid \boldsymbol{\omega})$ of the true class 
$k$ via RiemannianAdam \cite{rem_adaw}.

\subsubsection{Syntax Probing}
\label{syntax_probing}
Similarly, in this subsection, we obtain syntax through probing. The difference is that for syntax, we focus on tokens. In the projected hyperbolic space, the distance of the token embeddings from the origin and the distance between tokens correspond to the depth of the tokens and their distance in the DPT respectively.
We project word embeddings first:
\begin{equation}
    \boldsymbol{e}_{j}^{sy}=\boldsymbol{A}_{sy}\otimes_c\boldsymbol{e}_j,
\end{equation}
where $\boldsymbol{e}_j=\boldsymbol{E}_{j,:}$. How to parameterize a dependency tree from dense embeddings is non-trivial. Following \citet{structural_probe}, we define two metrics to measure the deviation from the standard: using the distance between two words in embedding space to represent the distance of word nodes in the dependency tree, and using the distance of a word from the origin to represent the depth of the word node. We use the following two loss functions:
\begin{equation}
    \mathcal{L}_{\text{dis}}=\frac{1}{n^2}\sum\limits_{j,j^{\prime}\in [n]}{|d_{DPT}(x_j,x_{j^{\prime}})-d_{\mathcal{B}}(\boldsymbol{e}_{j}^{sy},\boldsymbol{e}_{j^{\prime}}^{sy})^2|},
\end{equation}
\begin{equation}
    \mathcal{L}_{\text{dep}}=\frac{1}{n}\sum\limits_{j\in [n]}{|d_{DPT}(x_j)-d_{\mathcal{B}}(\boldsymbol{e }_{j}^{sy},\boldsymbol{0})^2|}.
\end{equation}
where $d_{DPT}(x_j,x_{j^\prime})$ and $d_{DPT}(x_j)$ represent the distance of words and the depth of words respectively. And $d_{\mathcal{B}}(\boldsymbol{e }_{j}^{sy},\boldsymbol{0})$ denotes the distance between $\boldsymbol{e}_{j}^{sy}$ and the origin in the projected hyperbolic space.
\subsection{Shapley Contribution Estimation}
\label{feature_contribution}
According to cooperative game theory, we regard the input as a set of players $N$, where each element of the set corresponds to a word, and the process of hierarchical clustering is viewed as a game, with clusters containing more than two words considered a coalition. Following \citet{zhangquanshi}, we define the characteristic function as $v=f$.
Given a game $\Gamma=(N,v)$, a fair payment scheme rewards each player according to its contribution. The Shapley value removes the dependence on ordering by taking the average over all possible orderings for fairness. The Shapley value of player $j$ in a game is as follows:
\begin{equation}
    \phi_j=\frac{1}{|N|!}\sum\limits_{\pi\in \Pi(N)}[v(Q_j^{\pi}\cup \{j\})-v(Q_j^{\pi})],
\end{equation}
where $\Pi(N)$ is the set of all permutations of the players, $Q_j^{\pi}$ is the set of players preceding player 
$j$ (i.e. token $j$) in permutation $\pi$. $v(S)$ is the value that the coalition of players $S\subseteq N$ can achieve together.
In practical, Monte Carlo sampling is used:
\begin{equation}
    \hat{\phi_j}=\frac{1}{R}\sum\limits_{r=1}^{R}{v(Q_j^{\pi_r}\cup \{j\})-v(Q_j^{\pi_r})}
\end{equation}
where $\pi_r$ denotes the $r$-th permutation in $\Pi(N)$.
Unfortunately,  Monte Carlo sampling methods can exhibit slow convergence \cite{mc_shapley}.

It is noticeable that attention mechanism of Transformer is  permutation invariant \cite{transformer, tranformer_order}, and the sinusoidal position embedding is only related to the specific position, not to the word. 
Moreover, after being trained with a Language Modeling task, the model has the ability to fill in the blanks based on context. 
Therefore, we assume that it is unnecessary to enumerate exponential combinations of words and the contribution of preceding permutation set (e.g.$\pi(<r)$) is included in larger subsequent permutation sets (e.g.$\pi(r)$). Therefore, we directly calculate contribution as follows:
\begin{equation}
    \label{context}
    \begin{split}
    &\tilde{\phi}_j=v(N)-v(N\setminus\{j\})\\
    &=f(X)-f(X\setminus\{x_j\})
    \end{split}
\end{equation}
where $N\setminus\{j\}$ denotes the player set excluding player $j$ and $X\setminus\{x_j\}$ denotes the input excluding token $x_j$. 
\subsection{Minimum Spanning Tree}
\label{minimum_spanning_tree}
\begin{figure}[!tb]
    \centering
    \includegraphics[width=0.5\textwidth]{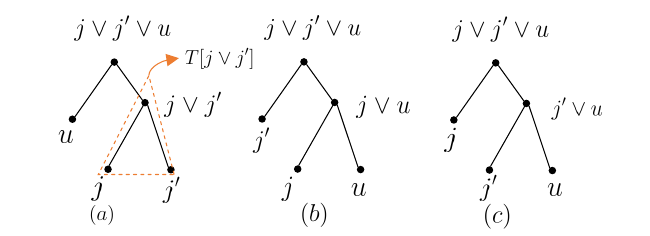}
    \caption{Three different binary tree types rooted from $j\vee j^\prime\vee u$.}
    \label{fig:model2}
\end{figure}

Our goal is to identify a hierarchy tree $T$ that aligns with semantic similarities, syntax similarities, and the contributions of individual elements. Building upon  \citet{Dasgupta}, we use the following cost:
\begin{equation}
    C_{D}(T;e)=\sum_{j,j^\prime\in [n]}e_{jj^\prime}|leaves(T[j\vee j^\prime])|,
\end{equation}
where $e_{j,j^\prime}$ denotes the pairwise similarities, $leaves(T[j\vee j^\prime])$ is leaves of $T[j\vee j^\prime]$, which is the subtree rooted at $j\vee j^\prime$, $j\vee j^\prime$ is the parent node of $j$ and $j^\prime$ as shown in Figure \ref{fig:model2}. 
Due to the unfolding dilemma of $
leaves(T[i\vee j])$ process, we adopt following expansion by \citet{wang2018improved}:
\begin{equation}
    \label{cost}
    \begin{split}
    C_{D}(T;e) & =\sum\limits_{jj^\prime u\in [n]}[e_{jj^\prime}+e_{ju}+e_{j^\prime u} \\
    & \quad -e_{jj^\prime u}(T)]+2\sum\limits_{jj^\prime}e_{jj^\prime},
\end{split}
\end{equation}
where 
\begin{equation}
    \begin{split}
        e_{jj^\prime u}(T)= & e_{jj^\prime}\mathbbm{1}[\{j,j^\prime\mid u\}]+e_{ju}\mathbbm{1}[\{j,u\mid j^\prime\}] \\ 
    & \quad +e_{j^\prime u}\mathbbm{1}[\{j^\prime,u\mid j\}],
    \end{split}
\end{equation}
where $\{j,j^\prime\mid u\}$ means the $j\vee j^\prime$ is the descendant of $j\vee j^\prime\vee u$, illustrated in Figure \ref{fig:model2}. The same for $\{j,u\mid j^\prime\}$ and $\{j^\prime,u\mid j\}$.

We aim to find the binary tree $T^*$:
\begin{equation}
    T^* = \mathop{\text{argmin}}_{\text{all binary trees } T} C_{D}(T;e).
\end{equation}
Directly optimizing this cost presents a combinatorial optimization problem. We introduce the following decomposition:
\begin{equation}
    \label{eq13}
    \begin{split}
        e_{jj^\prime} = &-\tilde{\phi}(j\vee j^\prime) + \alpha_1d_{\mathcal{B}}(\boldsymbol{e}_{j}^{se},\boldsymbol{e}_{j^\prime}^{se}) \\
         &+ \frac{1}{2}\alpha_2(d_{\mathcal{B}}(\boldsymbol{e}_{j}^{sy},\boldsymbol{0}) + d_{\mathcal{B}}(\boldsymbol{e}_{j^\prime}^{sy},\boldsymbol{0})),
    \end{split}
\end{equation}
where $\alpha_1\text{,} \alpha_2\in [0,1]$.

Under that we prove the optimal tree $T^{*}$ is a like-minimum spanning tree of Equation\ref{cost}.\footnote{The difference from the original minimum spanning tree is located in the last paragraph of Appendix \ref{sec:proof}.} 
The proof can be found in Appendix \ref{sec:proof}.
Ultimately we introduce the following decoding algorithm: 
\begin{algorithm}[!htbp]
    \caption{Building Algorithm}
    \label{alg2}
    \begin{algorithmic}[1]
        \REQUIRE Label hyperbolic embeddings $\boldsymbol{E}^{se}=\{\boldsymbol{E}_{1}^{se},\cdots,\boldsymbol{E}_{n}^{se}\}$, syntax hyperbolic embeddings $\boldsymbol{E}^{sy}=\{\boldsymbol{E}_{1}^{sy},\cdots,\boldsymbol{E}_{n}^{sy}\}$ 
        \ENSURE Binary tree $T$ with $n$ leafs
        \STATE $T\leftarrow (\{x_j\}:x_j\in X)$
        \STATE Initialize a PriorityQueue $\Upsilon$
        \STATE $\Upsilon\leftarrow \{(x_j,x_{j^\prime}):\text{pairs sorted by}\,e_{jj^\prime}\}$
        \WHILE{$\Upsilon\neq \varnothing$} 
            
            \STATE $x_j,x_{j^\prime}\leftarrow \Upsilon.$front, $\Upsilon.$pop
            \IF{$x_j$ and $x_{j^\prime}$ not in $T$} 
            \STATE $T\leftarrow T\cup \{x_j\vee x_{j^\prime} \}$
            \STATE $\Upsilon.$push($x_i\vee x_j$)
            
            \ENDIF
        \ENDWHILE
        
    \end{algorithmic}
    
\end{algorithm}

\section{Experiments}
 
\subsection{Experimental Setups}
\textbf{Datasets}\quad To evaluate the effectiveness of PE, we perform comprehensive experiments on three representative text classification datasets: \enquote{Rotten Tomatoes} \cite{pang-lee-2005-seeing}, \enquote{TREC} \cite{li-roth-2002-learning}, \enquote{Yelp} \cite{NIPS2015_250cf8b5}. Detailed statistics are in Table \ref{datasettable}.
\begingroup
\setlength{\tabcolsep}{4pt} 
\begin{table}[h!]
\begin{tabular}{llll}
\Xhline{1pt}
\textbf{Datasets}         & \textbf{Train/Dev/Test} & \textbf{C}     & \textbf{L}   \\
\Xhline{1pt}

Rotten Tomatoes & 10K/2K/2K     & 2 
& 64 \\
\Xhline{1pt}
TREC            & 5000/452/500         & 6 
& 64 \\
\Xhline{1pt}
Yelp            & 10K/2K/1K    &  2 
& 256 \\
\Xhline{1pt}
\end{tabular}
\caption{Statistics of three datasets. C: number of classes, L: average text length}
\label{datasettable}
\end{table}

\noindent \textbf{Metrics}\quad Following prior literature \cite{AOPC}, we use AOPC metric, which is the average difference of the change in predicted class probability before and after removing top $K$ words. 
\begin{equation}
    \text{AOPC}=\frac{1}{n}\sum_{K}(f(x_i)-f(\Tilde{x}_i^K))
\end{equation}
Higher is better.  And we evaluate two different strategies: $del$ and $pad$.
Concretely, We assign values to words through the following formula:
\begin{equation}
\label{aopc}
    \text{score}_i=\Tilde{\phi}(j)-\beta_1d_{\mathcal{B}}(\boldsymbol{e}_{j}^{se},\boldsymbol{c}_{k})-\beta_2d_{\mathcal{B}}(\boldsymbol{e}_{j}^{sy},\boldsymbol{0}),    
\end{equation}
where $\boldsymbol{c}_k$ is the prototype of predicted label $k$ in the semantic hyperbolic space, $\boldsymbol{0}$ is the origin in the syntatic hyperbolic space, $\beta_1, \beta_2\in [0,1]$.

\noindent \textbf{Infrastructures} All experiments are processed on one 15 core 2.6GHz CPU (Intel(R) Xeon(R) Platinum 8358P) and one RTX3090 GPU.

\noindent \textbf{Baselines}\quad  We compare PE with three hierarchical attribution methods: HEDGE \cite{chen-etal-2020-generating-hierarchical}, $\text{HE}_{LIME}$ , $\text{HE}_{LOO}$ \cite{he} and three feature interaction methods: SOC \cite{ig_v1}, Bivariate Shapley (BS)\cite{bivare_shapley} and ASIV \cite{asiv}.

\begingroup
\setlength{\tabcolsep}{4pt} 

\begin{table*}[!htbp]
\centering
\small
\begin{tabular}{lllllllllllll}
\toprule[1pt]
\multirow{3}{*}{\makecell[l]{\textbf{Datasets}\\\textbf{Methods}}}      & \multicolumn{6}{c}{\textbf{Rotten Tomatoes}}  & \multicolumn{6}{c}{\textbf{TREC}}  \\
&\multicolumn{3}{c}{$\text{AOPC}_{del}$} & \multicolumn{3}{c}{$\text{AOPC}_{pad}$} &\multicolumn{3}{c}{$\text{AOPC}_{del}$} & \multicolumn{3}{c}{$\text{AOPC}_{pad}$} \\
             &  $10\%$      & $20\%$  & Avg  &    $10\%$ & $20\%$ & Avg
            &   $10\%$     &  $20\%$ & Avg   &  $10\%$  & $20\%$& Avg \\
             \Xhline{1pt}
SOC    &   0.102   &  0.117  & $0.110_{\pm0.003}$   & 0.149    &   0.153  & $0.151_{\pm0.002}$   &     0.074     &0.087 & $0.081_{\pm0.001}$ & 0.097 & 0.099 & $0.098_{\pm0.001}$         \\
HEDGE      &  0.087    & 0.134  & $0.111_{\pm0.011}$     &  0.084   & 0.194  &$0.139_{\pm0.009}$     & 0.068         & 0.079& $0.074_{\pm0.004}$ & 0.095& 0.101 &$0.098_{\pm0.008}$       \\
$\text{HE}_{LIME}$ & 0.075     &   0.195 &$0.135_{\pm0.005}$    &  0.076   &  0.193&$0.135_{\pm0.009}$      &    0.063      & 0.072&$0.068_{\pm0.003}$ &  0.059 & 0.066&$0.063_{\pm0.007}$     \\
$\text{HE}_{LOO}$ &  0.062    &  0.117&$0.090_{\pm0.004}$   & 0.061 & 0.119 &$0.090_{\pm0.004}$   &       0.081   & 0.092 &$0.087_{\pm0.001}$ & 0.075 & 0.086 &$0.081_{\pm0.005}$    \\
 BS         &  0.109    &  0.121  &$0.116_{\pm0.013}$    &   0.103  &  0.185   &$0.144_{\pm0.009}$  &   0.099       & 0.104 & $0.102_{\pm0.003}$& 0.097&0.105 &$0.101_{\pm0.005}$\\
ASIV      & 0.101     &   0.113 &$0.107_{\pm0.005}$    & 0.098    &  0.181&$0.140_{\pm0.008}$      &   0.093       & 0.106&$0.199_{\pm0.006}$ &0.092  &0.113  &$0.103_{\pm0.003}$      \\
PE      &  \textbf{0.304}    &   \textbf{0.352} &$\textbf{0.328}_{\pm0.011}$    & \textbf{0.364}    &  \textbf{0.313} &$\textbf{0.339}_{\pm0.003}$     &     \textbf{0.214}    & \textbf{0.220} &$\textbf{0.217}_{\pm0.007}$&  \textbf{0.183}  & \textbf{0.174} &$\textbf{0.179}_{\pm0.004}$\\ 

\toprule[1pt]
\end{tabular}
\caption{AOPC comparison results of PE  with baselines on the Rotten Tomatoes and TREC dataset.}
\label{rt_trec_table}
\end{table*}
\begingroup
\setlength{\tabcolsep}{4pt} 
\begin{table}[!htbp]
\small
\begin{tabular}{llllll}
\toprule[1pt]
\textbf{Datasets}& \multicolumn{4}{c}{\textbf{Yelp}}  \\
\multirow{3}{*}{\textbf{Methods}}  & \multicolumn{2}{c}{$\text{AOPC}_{del}$} & \multicolumn{2}{c}{$\text{AOPC}_{pad}$}  & \multirow{2}{*}{\quad{$\overline{\boldsymbol{t}}$}} \\
&  $10\%$       & $20\%$    &    $10\%$ & $20\%$ \\
             \Xhline{1pt}
HEDGE      & 0.077  &  0.084     &  0.074   &  0.089 &  $70.312_{\pm0.074}$     \\
$\text{HE}_{LIME}$ &  0.056  &    0.075    &  0.065   &  0.076    &    $20.383_{\pm0.054}$\\
$\text{HE}_{LOO}$ &  0.040  &   0.071     & 0.059    &  0.064     & $16.201_{\pm0.079}$\\   
PE        & \textbf{0.110}   &   \textbf{0.138}     &  \textbf{0.112}   & \textbf{0.143}  &  $\textbf{2.230}_{\pm0.042}$     \\ 

\toprule[1pt]
    \end{tabular}
    \caption{AOPC and time efficiency comparision results of PE and baselines on the Yelp dataset. $\overline{\boldsymbol{t}}$ denotes the average time of building HA tree per input in seconds. }
    \label{yelp_table}
\end{table}

\subsection{General Experimental Results}
We first evaluate our method using the AOPC metric across three datasets, as shown in Tables \ref{rt_trec_table} and \ref{yelp_table}. \textbf{Firstly}, our method, PE, consistently surpasses the baseline in binary and multiclass tasks for both short and long texts. For instance, PE outperforms $\text{HE}_{LOO}$ by 0.235 in Table \ref{rt_trec_table} and by 0.067 in Table \ref{yelp_table} of $\text{AOPC}_{del}$,$20\%$, Rotten Tomatoes / Yelp setting.
\textbf{Second}, in comparison to recent works such as SOC and $\text{HE}_{LOO}$, our method's primary advantage lies in its computation efficiency. We conduct an analysis comparing the average time  of various approaches to construct HA trees. The  results in Table \ref{yelp_table} indicate that PE substantially outperforms its counterparts in terms of speed, being twice as fast as SOC and six times faster than $\text{HE}_{LIME}$. 

\subsection{Ablation Study}
We conduct ablation experiments with three modified baselines from PE: PE w/o prob corresponding $\tilde{\phi}(i)=0$, PE w/o semantic corresponding $\beta_1=0$ and PE w/o syntax corresponding $\beta_2=0$.

As shown in Figure \ref{fig:exp1}, both PE and variants outperform w/o prob baselines, demonstrating our approach’s effectiveness in directly calculating contributions in Equation \ref{context}. Moreover, we observe that both in $del$ and $pad$ settings, the utility of estimating contribution is more striking than the other two components in Equation \ref{aopc}. The reason may be that context has a greater impact on output than single semantics and syntax. It is noticeable that syntax slightly outperforms semantics, we hypothesis that the reason might be related to the nature of the tasks in the TREC dataset, as the labels tend to associate with syntactic structures \cite{li-roth-2002-learning}.

\begin{figure}[!tb]
    \centering
    \includegraphics[width=0.5\textwidth]{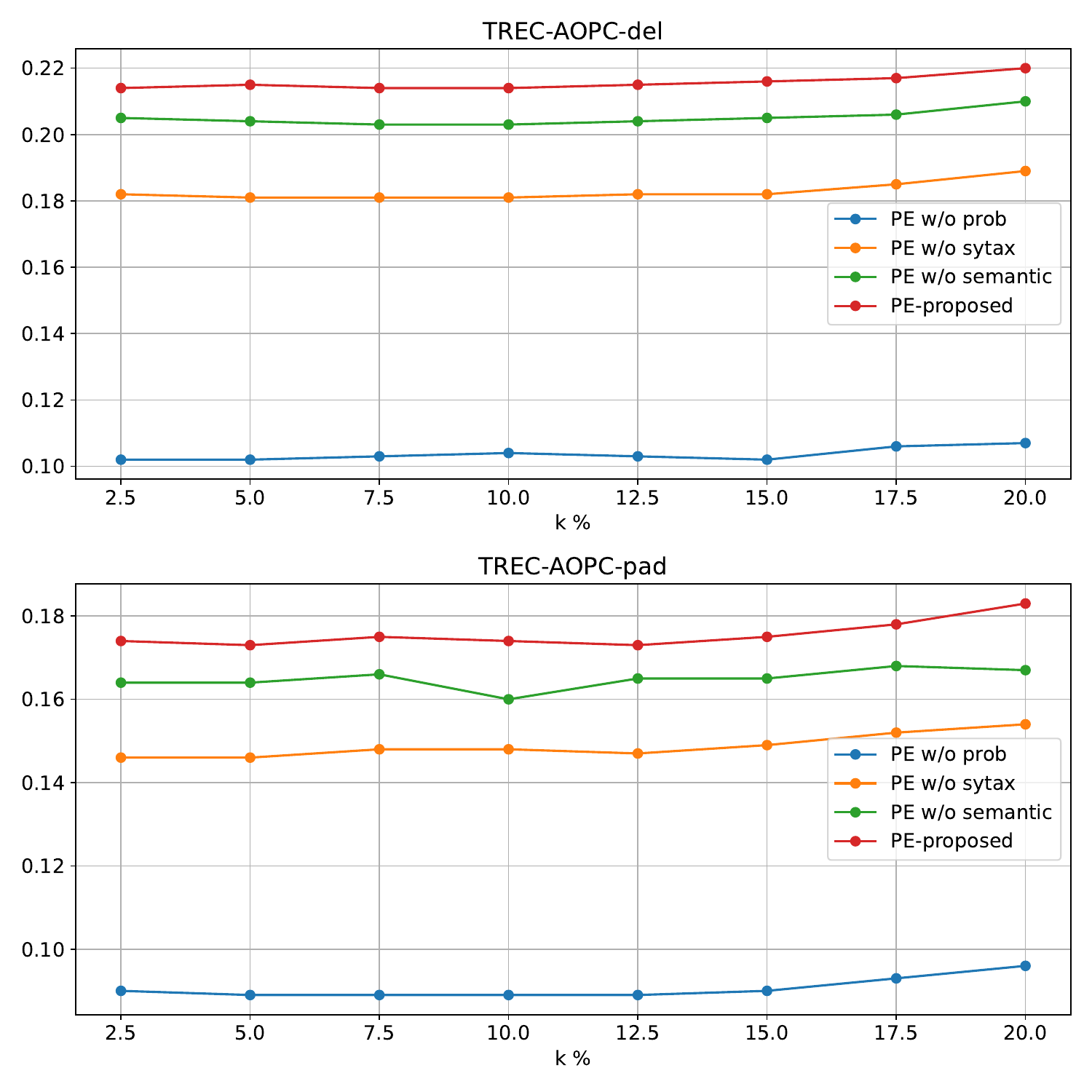}
    \caption{Evaluation results of Ablation Study.}
    \label{fig:exp1}
\end{figure}

\subsection{Case Study}
For qualitative analysis, we present two typical examples from the Rotten Tomatoes dataset to illustrate the role of PE in modeling the interaction of discontinuous features and we show more examples in Appendix \ref{sec:vis}. In the first example, we compare the results of PE and $\text{HE}_{LOO}$ in interpreting  BERT model. Figure \ref{fig:exp2} provides two hierarchical explanation examples for a positive and  negative review, each generated by PE and $\text{HE}_{LOO}$ respectively. In Figure \ref{fig:sub1}, it can be seen that PE accurately captures the combination of words with positive sentiment polarity: \emph{delightful}, \emph{out}, and \emph{humor}, and captures the key combination of \emph{out} and \emph{humor} at step 1. Additionally, this example includes a word with negative polarity: \emph{stereotypes}, where it can be observed that $\text{HE}_{LOO}$ captures its combination with \emph{in} and \emph{delightful}, missing the combination with \emph{out} and \emph{humor}.
In Figure \ref{fig:sub2}, PE captures the combination of \emph{slightest} and \emph{wit} in the first phase and complements it with the combination of \emph{lacking} at step 2. HE captures the combination of \emph{combination} and \emph{animation} at step 1, and it adds \emph{lacking} at step 2. We can infer that PE is able to capture the feature combination more related to the label at a shallow level, which demonstrates the effectiveness of our method.

Additionally, to more vividly demonstrate the role of semantics and syntax in building hierarchical explanations, we illustrate with two examples from the TREC dataset. As shown in Figure \ref{fig:sub3}, when $\alpha_2=0.5$, at the  level $L_3$, PE combines \emph{center}, \emph{temperature}, \emph{the}, \emph{earth} together. However, when $\alpha_2=0$, PE combines \emph{the}, \emph{temperature}, \emph{the}, \emph{earth} together. In the dependency parse tree of the sentence \emph{what is the temperature of the center of the earth}, \emph{the} distance to root is greater than \emph{center}. This indicates that incorporating syntactic information is meaningful for constructing convincing hierarchical explanations.
\begin{figure}[!htb]
    
    \subfigure[A positive example \enquote{\emph{My big fat greek wedding uses stereotypes in a delightful blend of sweet romance and lovingly dished out humor.}}]{
        \begin{minipage}{0.5\textwidth}
            \centering
            \includegraphics[width=\textwidth]{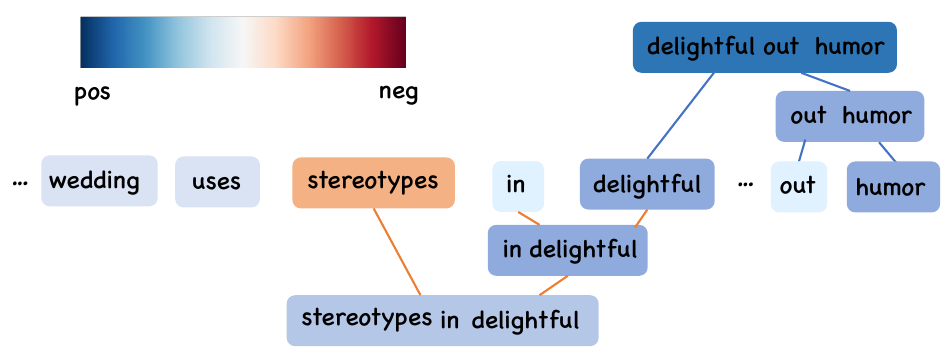} 
            \label{fig:sub1}
        \end{minipage}
    }
    \subfigure[A negative example \enquote{\emph{Just another combination of bad animation and mindless violence lacking the slightest bit of wit or charm.}}]{
        \begin{minipage}{0.5\textwidth}
            \centering
            \includegraphics[width=\textwidth]{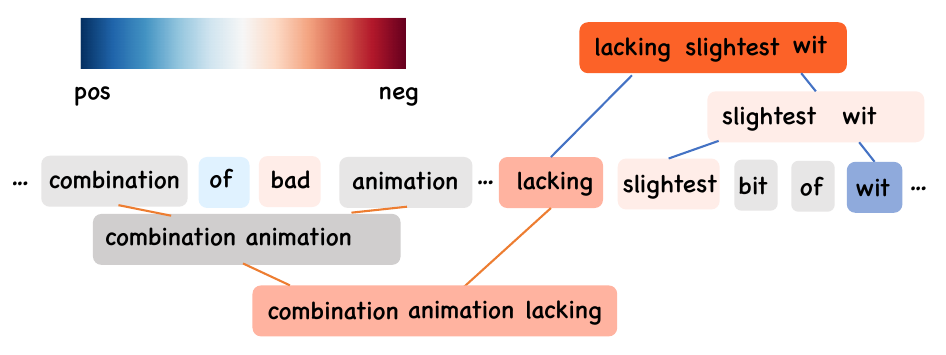}
            \label{fig:sub2}
        \end{minipage}
    } 
        \caption{PE,$\text{HE}_{LOO}$ for BERT on two examples from the Rotten Tomatoes dataset. The subtree in the upper right corner is generated by PE and the lower is produced by $\text{HE}_{LOO}$.}
\label{fig:exp2}
\end{figure}

\begin{figure}[!htb]
    
    \subfigure[An  example \enquote{\emph{What is the temperature at the center of the earth?}}, which the predicted label is numeric value.]{
        \begin{minipage}{0.5\textwidth}
            \centering
            \includegraphics[width=\textwidth]{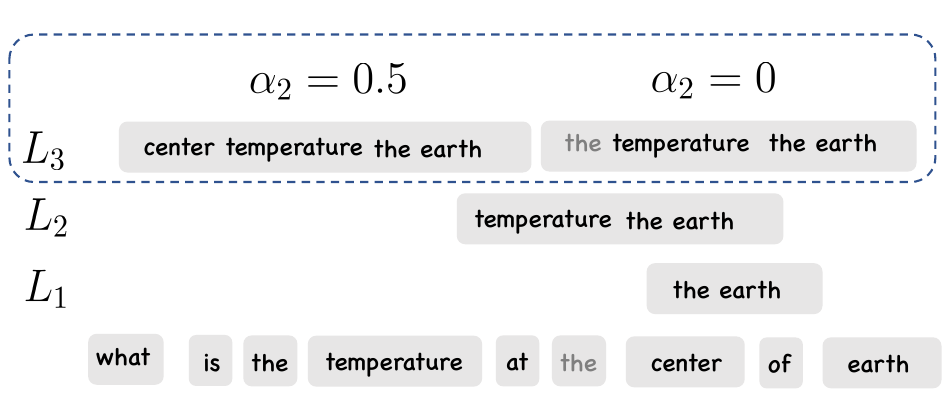} 
            \label{fig:sub3}
        \end{minipage}
    }
    \subfigure[A dependency parsing tree generated by Spacy \cite{spacy2}.]{
        \begin{minipage}{0.5\textwidth}
            \centering
            \includegraphics[width=\textwidth]{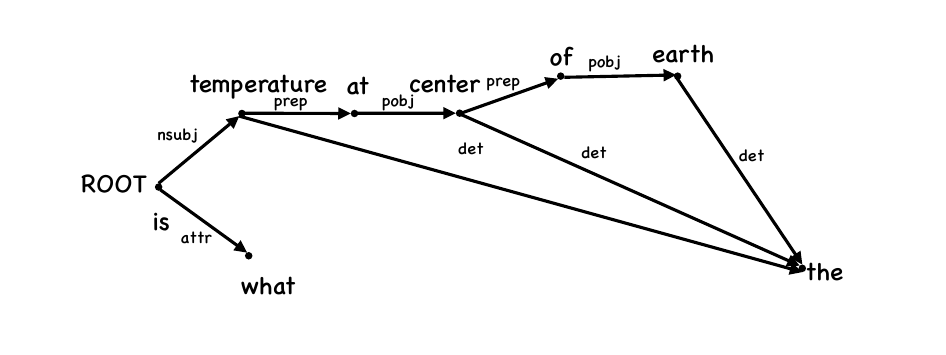}
            \label{fig:sub4}
        \end{minipage}
    } 
        \caption{PE for BERT on the example from the TREC dataset. The cluster on the left side of the third level $L_3$ is the  results for $\alpha_2=0.5$, and the right side is the result for $\alpha_2=0$.}
\label{fig:exp3}

\end{figure}

\section{Analysis of Time Complexity}
\label{discussion}
In this section, we delve into the time complexity associated with HA methods, which can be divided into two parts: the complexity of generating attribution scores, denoted as $\textbf{O}_{attr}$, and the complexity of generating the hierarchy from the scores, denoted as $\textbf{O}_{hierarchy}$. As shown in Table \ref{table:time_complexity}, we elaborate on the time complexity of various methods. For score computation, HEDGE utilizes the Monte Carlo sampling algorithm, with the number of samples denoted by $M_1$, leading to a time complexity of $O(nM_1)$. 
$\text{HE}_{LOO}$ uses the LOO algorithm \cite{LOO}, with a time complexity of $O(n^2M_1)$, where $M_2$ is the maximum number of iterations of the LOO algorithm. $\text{HE}_{LIME}$ method employs the LIME algorithm, with  ridge regression solving complexity of $O(n^3M_2)$, and $M_2$ is the number of sampled instances. The  time complexity of PE for solving scores is $O(n^2)$. 

\begingroup
\setlength{\tabcolsep}{2pt} 
\renewcommand{\arraystretch}{1.0}
\begin{table}[h!]
\centering
\begin{tabular}{lll}
\Xhline{1pt}
\textbf{Methods}         & $\textbf{O}_{attr}$  & $\textbf{O}_{hierarchy}$    \\
\Xhline{1pt}
HEDGE \citeyearpar{chen-etal-2020-generating-hierarchical}            &  $O(nM_1)$  &  $O(n^3)$ \\
\Xhline{1pt}
$\text{HE}_{LOO}$   \citeyearpar{he}      & $O(n^2M_2)$ & $O(n^3)$\\
\Xhline{1pt}
$\text{HE}_{LIME}$   \citeyearpar{he}     & $O(n^3M_3)$ & $O(n^3)$\\
\Xhline{1pt}
PE (ours) & $O(n^2)$ & $O(n^2logn)$ \\
\Xhline{1pt}
\end{tabular}
\caption{Comparison results of HA methods about capturing non-contiguous interactions and their time complexity. The relationship between the number of samples in the table and the value of \( n \) is: $M_1\gg M_2>M_3\gg n$.}
\label{table:time_complexity}
\end{table}

\section{Conclusion}
In this paper, we introduce PE, a computationally efficient method employing hyperbolic geometry for modeling feature interactions. More concretely, we use two hyperbolic projection matrices to embed the semantic and syntax information and devise a simple strategy to estimate the contributions of feature groups. Finally we design an algorithm to decode the hierarchical tree in an $O(n^2logn)$ time complexity. Based on the experimental results of three typical text classification datasets, we demonstrate the effectiveness of our method.

\section{Limitations}
The limitations of our work include:
1) Although our method boasts low time complexity, the use of the probing method to train additional model parameters, including two Poincare projection matrices, somewhat limits the generalizability of our approach.
2) In our experiments, we decompose the weights of the edges of the HA tree according to Equation \ref{eq13}. Whether there exists a  optimal decomposition formula remains  for future investigation.

\bibliography{emnlp2023-latex/custom}
\bibliographystyle{acl_natbib}

\appendix
\clearpage
\section{Proof}
\label{sec:proof}
First, we prove that the conclusion holds for $n=3$, and we generalize to the case of $n>3$ using induction. 

\noindent\textbf{\emph{Notation}} Due to the specificity of the binary tree we are solving for, a unique candidate tree can correspond to a node permutation $\pi$. For a tree with $n$ leaves, we define $\pi_n$ as the corresponding permutation.

We denote the constructed permutation  $\pi_n^*$ and prefix permutation $\pi_m^*$ in Algorithm \ref{alg2}.

\noindent\textbf{\emph{Base Case}} We here start the discussion from the left case in Figure \ref{fig:appendix1}. The cost can be expanded into:
\begin{equation}
    \begin{split}
        C_{D}(\pi_3^*;e) & =\sum\limits_{ijk}(e_{ik}+e_{jk})+2\sum\limits_{ij}e_{ij} \\
                    & = \sum_{ijk}2e_{ij}+e_{ik}+e_{jk}
    \end{split}
\end{equation}
Notice that $e_{ij}$ is smallest among $e_{ij}$, $e_{ik}$, $e_{jk}$ and among $\{i,j\mid k\}$, $\{i,k\mid j\}$, $\{j,k\mid i\}$, only one will hold true. We can conclude that $\pi_3^*$ is the solution that minimizes the cost.

\noindent\textbf{\emph{Induction Step}}  
We assume that the tree corresponding to the permutation $\pi_m$ has the smallest cost. To prove that $\pi_{m+1}$ is also the smallest.
We use a proof by contradiction to demonstrate that $\pi_{m+1}$ corresponds to the tree with the smallest cost. We define the tree's level as $L_1,\cdots,L_{n-1}$ in Figure \ref{fig:appendix1}. Firstly, we introduce the following lemma: 

\noindent\textbf{{Lemma}} We denote the $\gamma$-th step permutation produced in Algorithm \ref{alg2} as $\pi_{\gamma}^*$, and its corresponding tree cost as $C(\pi_{\gamma}^*)$. Now, if we swap the nodes at  level $L_{s}$ and $L_{t}$, $s<t$, and the resulting sequence ${\pi_{\gamma}^*}^\prime$, then $C({\pi_{\gamma}^*}^\prime)>C(\pi_{\gamma}^*)$.
\begin{proof}\let\qed\relax
We consider the cost after the swap as three parts: the triples that do not include $s$ and $t$, the part of the triples that include $s$ and the part that include $t$, denoted as $C_1$,$C_2$ and $C_3$. For ease of proof, we denote the sequence to the left of $s$ as $A=\pi^{*}_{\gamma,1:s-1}$, and the sequence between $s$ and $t$ as $B=\pi^{*}_{\gamma,s+1:t-1}$.
Obviously $C_1$ remains unchanged, as for $C_2$, before and after the swap:
\begin{equation}
    C_2=\sum\limits_{i,j\in A,s}e(\cdot)+\sum\limits_{i\in A,s,j\in B}e(\cdot)+\sum\limits_{s,i,j\in B}e(\cdot),
\end{equation}
\begin{equation}
    C_2^\prime=\sum\limits_{i,j\in A,s}e(\cdot)+\sum\limits_{i\in A,j\in B,s}e(\cdot)+\sum\limits_{i,j\in B,s}e(\cdot)
\end{equation}
By subtracting, we obtain:
\begin{equation}
    \begin{split}
       C_2^\prime-C_2 &= (\sum\limits_{i\in A,j\in B,s}e(\cdot)-\sum\limits_{i\in A,s,j\in B}e(\cdot)) + \\ & (\sum\limits_{i,j\in B,s}e(\cdot)-\sum\limits_{s,i,j\in B}e(\cdot))\geq 0. 
    \end{split}
\end{equation}
Similarly we obtain:
\begin{equation}
    \begin{split}
       C_3^\prime-C_3 &= (\sum\limits_{i\in A,t,j\in B}e(\cdot)-\sum\limits_{i\in A,j\in B,t}e(\cdot)) + \\ & (\sum\limits_{t,i,j\in B}e(\cdot)-\sum\limits_{i,j\in B,t}e(\cdot))\geq 0.  
    \end{split}
\end{equation}
\end{proof}
Now we prove that $\pi_{m+1}$ is smallest. If $\pi_{m+1}$ is not the smallest, then the node at the last level can be the smallest by swapping with a previous node. There are two cases: when the swapped node is from the first level (e.g. $j$), in this case, the difference in cost before and after the swap becomes:
\begin{equation}
    \begin{split}
        \Delta C &= (\sum\limits_{i\in C,m+1,j \in D}e(\cdot)-\sum\limits_{i\in C,j\in D,m+1}e(\cdot)) + \\ & (\sum\limits_{t,i,j\in D}e(\cdot)-\sum\limits_{i,j\in D,t}e(\cdot))\geq 0,
    \end{split}
\end{equation}
where $C=\pi_{m+1,1}^*$, $D=\pi_{m+1,3:m}^*$.
Similarly, when the swapped node is located in other levels, the cost after the swap will not decrease. This means that in $C(\pi_{m+1})$ cannot be smaller through swapping other leaves from different levels, thus $\pi_{m+1}$ is smallest. 

\begin{figure}
    \centering
    \includegraphics[width=0.5\textwidth]{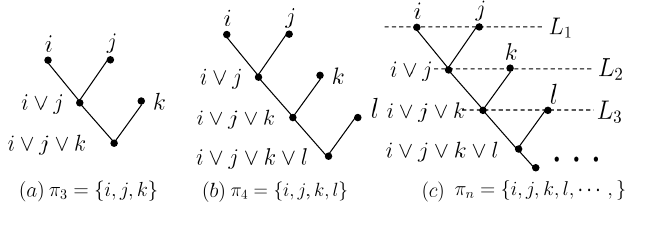}
    \caption{Examples for $\pi_3$, $\pi_4$ and $\pi_n$.}
    \label{fig:appendix1}
\end{figure}

The primary difference is that the edge weights in our graph \cite{mst} are not all known in advance but are dynamically generated. 

\section{Visualization}
\label{sec:vis}
\begin{figure*}[!htb]
    
    \subfigure[A negative example \enquote{\emph{The redeeming feature of Chan's films has always been the action, but the stunts in the tuxedo seem tired and what's worse, routine.}}]{
        \begin{minipage}{\textwidth}
            \centering
            \includegraphics[width=\textwidth]{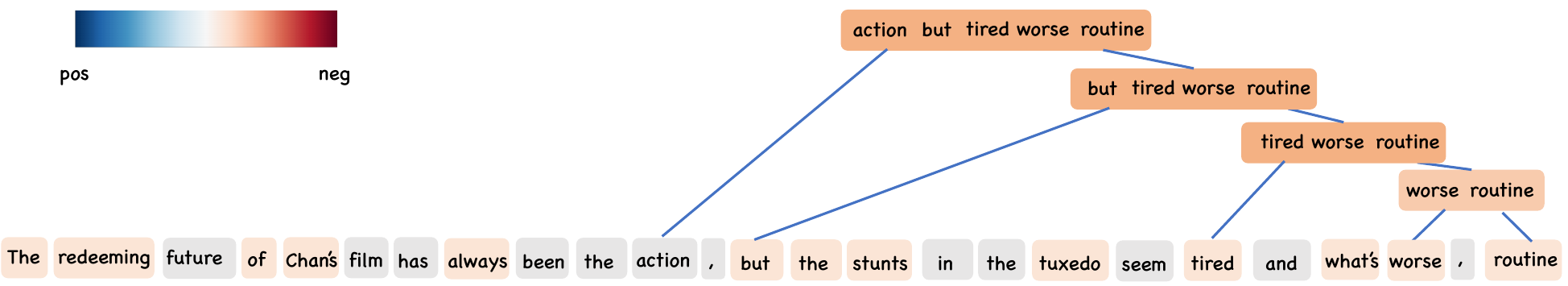} 
            \label{fig:appendix2-1}
        \end{minipage}
    }
    \subfigure[A positive example \enquote{\emph{The production values are of the highest and the performances attractive without being memorable.}}]{
        \begin{minipage}{\textwidth}
            \centering
            \includegraphics[width=\textwidth]{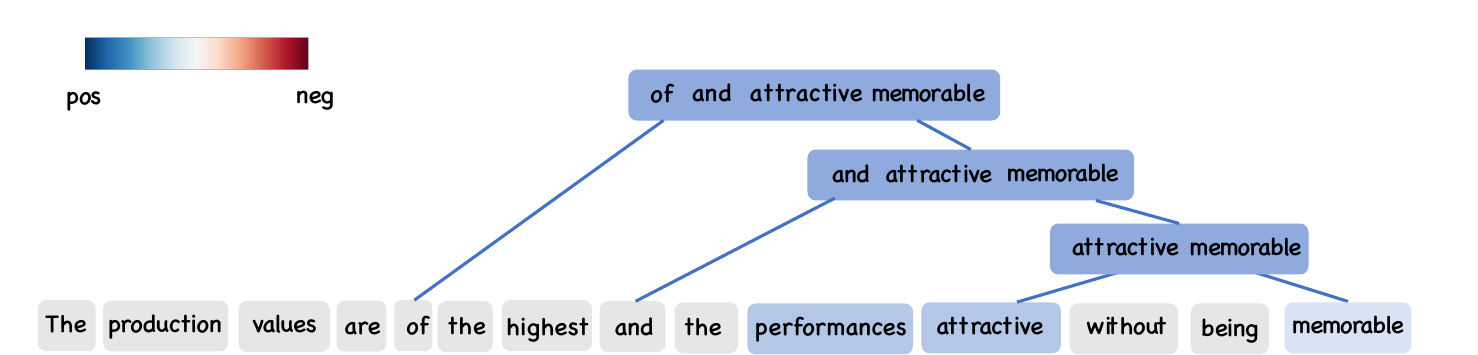}
            \label{fig:appendix2-2}
        \end{minipage}
    } 
        \caption{PE for BERT on two examples from the Rotten Tomatoes dataset.}
\label{fig:appendix3}

\end{figure*}

\begin{figure*}[!htbp]
    
    \subfigure[A negative example \enquote{\emph{Service here sucks \textbackslash n I love the food still \textbackslash n\textbackslash n but the service is so bad.}}]{
        \begin{minipage}{\textwidth}
            \centering
            \includegraphics[width=\textwidth]{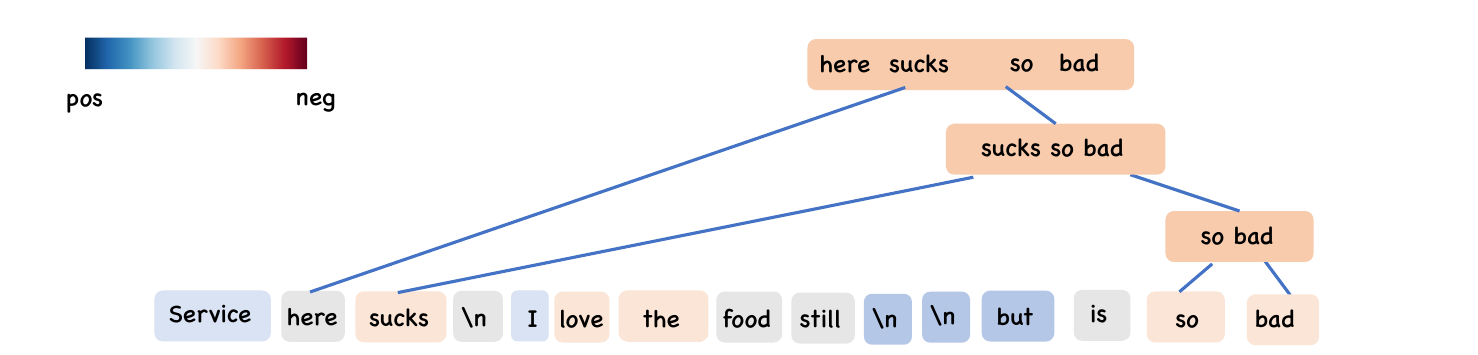} 
            \label{fig:appendix2-3}
        \end{minipage}
    }
    \subfigure[A positive example \enquote{\emph{Flavors are great but every time I come this location it is disgusting machines are dirty.}}]{
        \begin{minipage}{\textwidth}
            \centering
            \includegraphics[width=\textwidth]{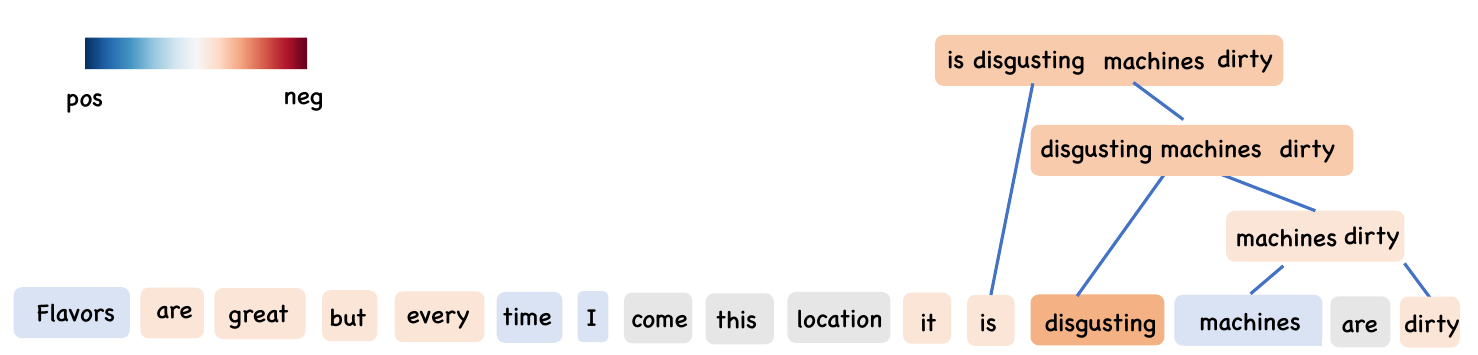}
            \label{fig:appendix2-4}
        \end{minipage}
    } 
        \caption{PE for BERT on two examples from Yelp dataset.}
\label{fig:appendix4}

\end{figure*}

\section{Implementation Details}
\label{sec::implementation_details}
In this work, all language models are implemented by Transformers. All our experiments are performed on one A800. The results are reported with 5 random seeds.

For fine tuning the projection matrix $\boldsymbol{P}_c$, we iterate 5 epochs using RiemanianAdam optimizer and learning rate is initialized as 1e-3, the batch size is 32.
For fine tuning the projection matrix $\boldsymbol{P}_s$, we use the Penn Treebank dataset we iterate 40 epochs using Adam optimizer and learning rate is initialized as 1e-3. We set $d_{out}$ as 64.
We use grid search to search $\alpha_1, \alpha_2, \beta_1, \beta_2\in \{0,0.1,0.2,0.3,0.4,0.5\}$.
\section{HA Example}
\label{HA example}
\begin{figure}[!htbp]
    \centering
    \includegraphics[width=0.51\textwidth]{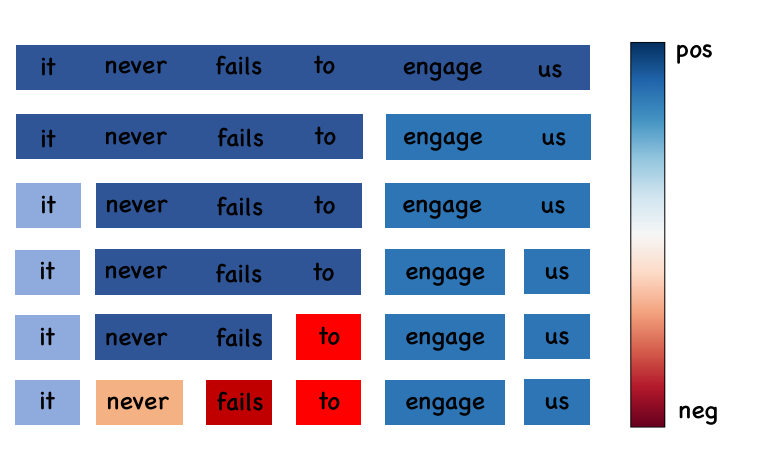}
    \caption{A hierarchy example from HEDGE \cite{chen-etal-2020-generating-hierarchical}. The background color of the words and phrases represents emotional polarity, with cool colors indicating positive and warm colors indicating negative.}
    \label{fig:appendix1}
\end{figure}
\section{Lime Explanation}
\label{lime_expl}
\begin{figure}[!htbp]
    \centering
    \includegraphics[width=0.45\textwidth]{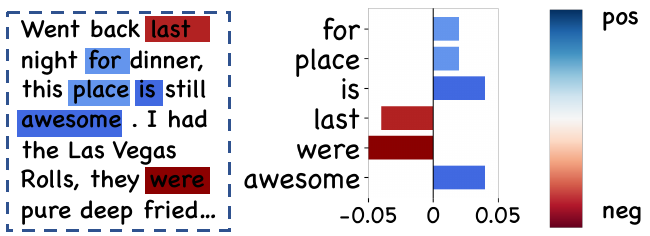}
    \caption{A LIME explanation example from a random forest classifier. It can be observed that two stop words (i.e.\enquote{is} and \enquote{were}) are identified as positive and negative emotional polarities, respectively.}
    \label{fig:intro1}
\end{figure}

\end{document}